\documentclass{article} 
\pdfoutput=1
\PassOptionsToPackage{numbers}{natbib}

\usepackage[final]{neurips_2024} 

\usepackage[utf8]{inputenc} 
\usepackage[T1]{fontenc}    
\usepackage{hyperref}       
\usepackage{url}            
\usepackage{soul}
\usepackage{adjustbox}
\usepackage{booktabs}       
\usepackage{amsfonts}       
\usepackage{nicefrac}       
\usepackage{microtype}      
\usepackage{xcolor}         
\usepackage{verbatim}
\usepackage{blindtext}
\usepackage{acronym}
\usepackage{graphicx}
\usepackage{subfigure}
\usepackage{comment}
\usepackage{placeins}
\usepackage{tikz}
\usepackage[ruled,linesnumbered,noend]{algorithm2e}
\usepackage{amssymb}
\usepackage{amsmath}
\usepackage{amssymb}
\usepackage{mathtools}

\usepackage{wrapfig}

\usepackage{amsthm}
\usepackage{wrapfig}
\usepackage[textsize=tiny,textwidth=2cm]{todonotes}
\usepackage{pgfplots}
\usepackage{floatrow} 
\usepackage{multirow}
\usepgfplotslibrary{fillbetween}
\usepgfplotslibrary{groupplots}
\usepackage{array}
\usepackage{floatrow} 
\usepackage{parskip}

\newtheorem{definition}{Definition}

\newtheorem{proposition}{Proposition}
\newtheorem{corollary}{Corollary}

\pgfplotsset{compat=1.18} 
\usetikzlibrary{intersections}
\usepgfplotslibrary{fillbetween}
\usepgfplotslibrary{groupplots}

\title{C-MCTS:\\Safe Planning with Monte Carlo Tree Search}

\author{%
  Dinesh Parthasarathy
  \\
  FAU Erlangen-N\"urnberg\\
  Erlangen, Germany \\
  \texttt{dinesh.parthasarathy@fau.de}
  \AND
  Georgios Kontes $~~$ $~~$ Axel Plinge $~~$ $~~$ Christopher Mutschler\\
  Fraunhofer Institute for Integrated Circuits (IIS), Fraunhofer IIS \\
  Nuremberg, Germany\\
  \texttt {\{FirstName.LastName\}@iis.fraunhofer.de} }

\acrodef{RL}{Reinforcement Learning}
\acrodef{MCTS}[MCTS]{Monte Carlo Tree Search}
\acrodef{MPC}[MPC]{Model Predictive Control}
\acrodef{CEM}[CEM]{Cross Entropy Method}
\acrodef{POMDP}[POMDP]{Partially Observable Markov Decision Process}
\acrodef{CPOMDP}[C(PO)MDP]{Constrained (Partially Observable) Markov Decision Process}
\acrodef{MDP}[MDP]{Markov Decision Process}
\acrodef{RMDP}[RMDP]{Robust Markov Decision Process}
\acrodef{MOMDP}[MOMDP]{Mixed Observable Markov Decision Process}
\acrodef{CMDP}[CMDP]{Constrained Markov Decision Process}
\acrodef{CC-POMCP}[CC-(PO)MCP]{Cost-Constrained (Partially Observable) Monte Carlo Planning}
\acrodef{CC-MCP}[CC-MCP]{Cost-Constrained Monte Carlo Planning}
\acrodef{TD}[TD]{Temporal Difference}
\acrodef{UCB}[UCB]{Upper Confidence Bound}
\acrodef{SARSA}[SARSA]{State–action–reward–state–action}
\acrodef{UCT}[UCT]{Upper Confidence bounds applied for Trees}
\acrodef{LP}{Linear Program}
\acrodef{OOD}{out-of-distribution}

\definecolor{darkgreen}{RGB}{0,100,0}

\begin{document}

\maketitle

\begin{abstract}%
The Constrained Markov Decision Process (CMDP) formulation allows to solve safety-critical decision making tasks that are subject to constraints. 
While CMDPs have been extensively studied in the Reinforcement Learning literature, little attention has been given to sampling-based planning algorithms such as Monte Carlo Tree Search (MCTS) for solving them.
Previous approaches are conservative with respect to costs as they avoid constraint violations by using Monte Carlo cost estimates that suffer from high variance.
We propose Constrained MCTS (C-MCTS), which estimates cost using a safety critic that is trained with Temporal Difference learning in an offline phase prior to agent deployment. The critic limits exploration to unsafe regions during deployment by pruning unsafe trajectories within MCTS. This makes C-MCTS more efficient w.r.t. planning steps. Compared to previous work, it achieves higher rewards by operating closer to the constraint boundary (while satisfying cost constraints) and is less susceptible to cost violations under model mismatch between the planner and the deployment environment.
\end{abstract}%

\section{Introduction}
\label{sec:introduction}

\ac{MCTS} is a decision-making algorithm that employs Monte Carlo methods across the decision space, evaluates their outcome with respect to a given reward/objective, and constructs a search tree focusing on the most promising sequences of decisions~\citep{mcts_survey,mcts_review}. The success of~\ac{MCTS} lies in the asymmetry of the trees constructed, which ensures better exploration of promising parts of the search space. The possibility of using neural networks as heuristics to guide the search tree has helped tackle complex and high-dimensional problems with large state and action spaces~\citep{alphazero}.

However, as vanilla~\ac{MCTS} only optimizes for a single objective it is unsuitable for a large class of real-world problems that also require a set of constraints to be fulfilled. These types of problems are usually modeled as Constrained Markov Decision Processes (CMDPs)~\citep{altman_CMDP} and specialized algorithms are used to solve them. Such algorithms include approaches that rely on expert knowledge to create safe action sets~\citep{Hoel,mohammadsafedecision,mirchevska2018high}, Lagrangian relaxation methods that update primal and dual variables incrementally online and learn safe policies~\citep{Ding_PPO_CMDP,dual_CMDP}, approaches that learn separate reward and cost/constraint signals to train a safe-aware policy both in~\ac{MDP}~\citep{safetycritic_bharadh,safetycritic_krishnan,safetycritic_yang} and~\ac{RMDP} environments~\citep{tamar2014scaling,mankowitz2020robust}, and methods that use uncertainty-aware estimators like Gaussian Processes to balance exploration-exploitation risk~\citep{Wachi2018,hewing2019cautious}. 

\begin{figure}[!t]
\centering
\includegraphics[width=\textwidth]{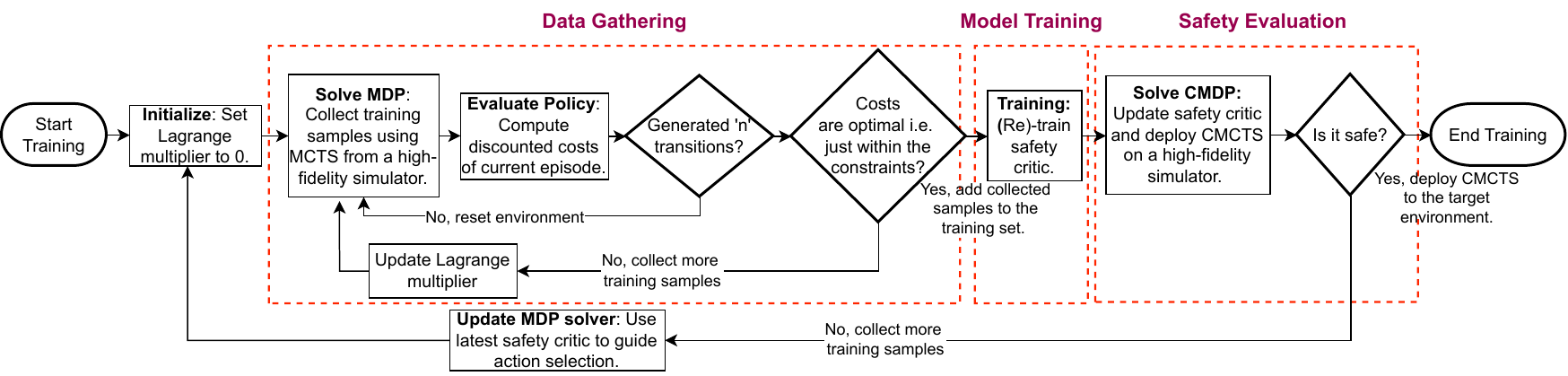}
\caption{Simplified flow of training phase in C-MCTS.}
\label{fig:train_part}
\end{figure}

We propose Constrained MCTS (C-MCTS), an~\ac{MCTS}-based approach for solving CMDPs (Fig.~\ref{fig:train_part}). We use a high-fidelity simulator to collect trajectories under different safety constraint satisfaction levels. This allows for violating cost-constraints during training with no impact, as well as simulating rare events that have a safety impact. With these samples, we train a safety critic \emph{offline}, which is used during deployment within \ac{MCTS} to make cost predictions and avoid tree expansion to unsafe states. C-MCTS constructs deeper search trees with fewer planning iterations compared to the state-of-the-art while operating safely closer to the cost-constraint, thus leading to higher rewards.

\section{Monte Carlo Tree Search for Constrained MDPs}
\label{sec:problem}

A~\ac{CMDP} can be defined by the tuple $\left <S,A,P,R, \mathbf{C},\mathbf{\hat{c}},\gamma,\mu \right>$ where $S$ is the set of states $s$, $A$ is the set of actions $a$, $P$ defines the probability of transitioning from $s \in S$ to $s' \in S$ for action $a \in A$ executed at $s$, $R$ is a reward function that returns a one-step reward for a given action $a$ at a state $s$, $\gamma \in [0,1)$ is the discount factor, and $\mu: S \mapsto \left[0, 1\right]$ is the initial state distribution. Following the notation convention of~\cite{CC-MCP}, $\mathbf{C} = \{C_m\}_{1\ldots M}$ is a set of $M$ non-negative cost functions, with $\mathbf{\hat{c}} = \{c_m\}_{1\ldots M} \in [0,1]$ their respective thresholds, \emph{in terms of average cost allowed per episode}. For the remainder of the text, we assume only one constraint function $C$ with its respective threshold $\hat{c}$, to simplify the notation. The optimal policy $\pi^* \in \Pi$ is a policy that belongs to a (parametric) policy class $\Pi$ that maximizes the expected discounted cumulative reward $V^\pi_R(s)$, while satisfying all the constraints on the expected discounted cumulative cost $V^\pi_C(s)$, as follows:
\begin{align}
   & \operatorname*{max}_{\pi \in \Pi} V_{R}^{\pi}(s) = \mathbb{E}_{\pi} \left[\sum_{t=0}^{\infty} \gamma^t R(s_t,a_t)|s_0=s\right]
   &s.t. \quad V_{C}^{\pi}(s) =  \mathbb{E}_{\pi} \left[\sum_{t=0}^{\infty} \gamma^t C(s_t,a_t)|s_0=s\right] \leq \hat{c}.
   \label{eq:constr_opt}
\end{align}
Depending on the context we will use the definitions of Eq.~\ref{eq:constr_opt} or the notion of the state-action expected discounted cumulative reward/cost (i.e., the state-action value function), defined (for cost) as follows: 
\begin{equation}
Q_C^\pi(s,a) = \mathbb{E}_{\pi} \left[\sum_{t=0}^{\infty} \gamma^t C(s_t,a_t)|s_0=s, a_0=a\right]  \triangleq V_{C}^{\pi}(s).
\label{eq:qc}
\end{equation}
Similar to assumptions of previous work (see e.g.,~\cite{tessler2018reward} and the robust constraint objective (Eq.~\ref{eq:qc}) from~\cite{mankowitz2020robust}), we prioritize the constraint satisfaction part of Eq.~\ref{eq:constr_opt}.
\begin{definition}
\citep{tessler2018reward} A feasible solution of the constrained optimization problem defined in Eq.~\ref{eq:constr_opt} is a solution that satisfies $V_{C}^{\pi}(s) \leq \hat{c}$.
\label{def:solution}
\end{definition}
One approach to address the problem in Eq.~\ref{eq:constr_opt} is using the Lagrange multiplier technique (see \cite{bertsekas2014constrained}). For formulating the Lagrangian, we can define the following:
\begin{definition}
\citep{tessler2018reward} The penalized reward function is defined as $r_\lambda(\lambda, s, a) = r(s,a) - \lambda\, c(s,a)$. The penalized expected discounted cumulative reward function is defined as $V_{R}^{\pi}(\lambda, s) = V_{R}^{\pi}(s)-\lambda V_{C}^{\pi}(s)$.%
\label{def:pen_reward}%
\end{definition}%
The Lagrangian transforms the posed problem into an unconstrained one:
\begin{equation}
   \operatorname*{min}_{\lambda \geq 0} \operatorname*{max}_{\pi \in \Pi} L(\lambda, \pi) = \operatorname*{min}_{\lambda \geq 0} \operatorname*{max}_{\pi \in \Pi} \left[ V_{R}^{\pi}(s) - \lambda \left(V_{C}^{\pi}(s) - \hat{c}\right)\right].
   \label{eq:lagrange}
\end{equation}
Even though a significant body of work in solving CMDPs is available, \ac{MCTS} for \emph{discrete-action}~\acp{CMDP} has only been little explored. 
To our knowledge, apart from the seminal work of~\cite{CC-MCP}, previous work extended~\ac{MCTS} only to multi-objective variants~\citep{Hayes2023} that attempt to construct local~\citep{ParetoMCTS} or global~\citep{mo-mcts} Pareto fronts and determine the Pareto-optimal solution. These approaches report good results at the expense of higher computational costs, due to the need to compute a set of Pareto-optimal solutions. Lee et al.~\cite{CC-MCP} proposed Cost-Constrained Partially Observable Monte Carlo Planning, an \ac{MCTS} algorithm to solve Constrained Partially Observable Markov Decision Process problems, which can be used to solve \ac{CMDP} settings (we will refer to this algorithm as \ac{CC-MCP}). \ac{CC-MCP} uses a Lagrange formulation and updates the Lagrange multiplier while constructing the search tree based on accumulated cost statistics. The \ac{CMDP} problem is formulated as a \ac{LP}, and then the dual formulation is solved: 
\begin{equation}
    \min_{\lambda \geq 0}\left[V_{R}^{*}(\lambda, s) + \lambda \hat{c}\right]
\label{eq:ccmcp_pwlinear}
\end{equation}
Here, $V_{R}^{*}(\lambda, s)$ is the optimal penalized expected discounted cumulative reward function, and $\hat{c}$ are the cost constraints. As the objective function in Eq.~\ref{eq:ccmcp_pwlinear} is piecewise-linear and convex over $\lambda$~\citep{CC-MCP}, $\lambda$ can be updated using the gradient information $V_{C}^{*}-\hat{c}$, where $V_{C}^*$ are the costs incurred for an optimal policy with a fixed $\lambda$. Hence, the~\ac{CMDP} can be solved by iterating the following three steps: (i) Solve MDP with a penalized reward function (see Definition~\ref{def:pen_reward}), (ii) evaluate $V_C^*$ for this policy, and (iii) update $\lambda$ using the gradient information. Steps (i) and (ii) can also be interleaved at a finer granularity, and this is the idea behind \ac{CC-MCP}, where $\lambda$ is updated at every \ac{MCTS} iteration based on the Monte Carlo cost estimate $\hat{V}_C$ at the root node of the search tree.

\section{Constrained Monte Carlo Tree Search (C-MCTS)}
\label{sec:method}

CC-MCP has several shortcomings: (1) it requires a large number of planning iterations to tune the Lagrange multiplier (as it is tuned online and thus CC-MCP explores both unsafe and safe trajectories in the search tree); (2) the agent acts conservatively, trying to satisfy the cost constraints; and (3) the algorithm also relies on the planning model to calculate cost estimates, making it error-prone due to the fast-but-inaccurate nature of planning models used for online planning at deployment.

In C-MCTS the training phase consists of approximating a safety critic that is used by the~\ac{MCTS} policy during the deployment phase (without a Lagrange multiplier) for pruning unsafe trajectories/sub-trees. We foresee the availability of two simulators: an inaccurate low-fidelity simulator, whose low complexity allows for utilization in the~\emph{online} planning/rollout phase of~\ac{MCTS}, and a high-fidelity one, used for data collection and evaluation of the safety critic training. We also assume that using the high-fidelity model for online planning during deployment is infeasible due to computational constraints, therefore we learn safety constraints in an offline phase using the high-fidelity simulator.


We train the safety critic offline (``Model Training'' in Fig.~\ref{fig:train_part}), by gathering samples from the training environment (high-fidelity simulator). As~\ac{MCTS} explores the state space exhaustively during \emph{online planning}, some state-action pairs are likely to be \emph{out-of-distribution}, i.e., some trajectories are not encountered during (offline) training. More formally, we have two main sources of inaccurate safety critic predictions: the \emph{aleatoric} and the \emph{epistemic} uncertainty. The former is inherent in the training data (e.g., due to the stochastic nature of the transition model) and the latter is due to the lack of training data (e.g., it could appear as an extrapolation error) -- see for example~\citep{chua2018deep} for a more formal discussion. To mitigate the effect of both uncertainty sources, we resort to a combination of utilizing an ensemble for the safety critic and selecting data near the constraint-switching hypersurface for the training phase. However, a large mismatch between the training simulator and the target environment, i.e., \emph{distribution shifts}, may still affect the agent's performance (Appendix \ref{app:robustness}).

\textbf{Uncertainty-aware safety predictions.} For the training, we use SARSA(0)~\citep{sutton2018reinforcement} (a~\ac{TD} Learning-like method~\citep{TD_sutton}), but instead of training a single safety critic, we train an ensemble. The individual members of the ensemble have the form of neural networks and approximate the state-action-cost function. We denote this ensemble safety critic as $\hat{Q}_{sc}^{*}(s,a)$. The trainable parameters of each member of the ensemble are optimized to minimize the mean-squared TD-error which uses a low variance one-step target. The aggregated ensemble output ($\hat{\mu},\hat{\sigma}$) provides a mean and a standard deviation computed from the individual member's outputs, which we then use within \ac{MCTS}. Hence, the safety critic output with an ensemble standard deviation greater than a set threshold $\hat{\sigma}>\sigma_{max}$ can be used to identify and ignore those samples and predictions.


The trained safety critic ensemble is used during the expansion phase in \ac{MCTS}, see Alg.~\ref{alg:c_mcts} -- the other phases (selection, simulation, backpropagation) are identical to vanilla \ac{MCTS}. At the expansion phase, we try to expand the search tree from the leaf node along different branches corresponding to different actions. First, based on the safety critic's output we filter out predictions that we cannot trust (corresponding to high ensemble variance) and create a reduced action set (lines 6-7). The safety of each action from this set is evaluated based on the safety critic's output predicting expected cumulative costs from the leaf. This is summed up with the one-step costs stored in the tree from the root node to the leaf node. If this total cost estimate is greater than the cost constraints ($\hat{c}$), then we prune the corresponding branches, while other branches are expanded (lines 8-11). These steps, when repeated over multiple planning iterations create a search tree exploring a safe search space.

\IncMargin{1.5em}
\begin{algorithm}[!t]
  \DontPrintSemicolon
  \caption{\textbf{C-MCTS} | Using a learned safety critic in \ac{MCTS}.}
  \label{alg:c_mcts}
  
  \( \mathcal{N}_{root} \) : Root node representing the current state, \( s_0 \).\\
  \( \mathcal{N}_{leaf} \) : Selected leaf node with state \( s_t \).\\
  \( \mathcal{P} \) : Traversed path from the root node to the leaf node \( (s_0,a_0,s_1,a_1,...,a_{t-1},s_t) \).\\
  
  \Repeat{maximum number of planning iterations is reached}{    
    $\mathcal{P}, \mathcal{N}_{leaf} \gets \textsc{Select}(\mathcal{N}_{root})$ \tcp{SELECTION (using UCT algorithm)}
    
    \tcp{EXPANSION}
    i. Get safety critic outputs $(\hat{\mu},\hat{\sigma})$ for all actions $a_t \in A$ from $\mathcal{N}_{leaf}$.\\
    ii. Identify feasible actions i.e. ${A}_\text{feasible} = \{a_t: \hat{\sigma}_{a_t} \leq \sigma_{max}\}$.\\
    iii. Calculate the cost estimate $\hat{Q}_{sc}^{*}(s_t,a_t)$ for actions $a_t \in {A}_\text{feasible}$. \\
    iv. Define: \( C_{path} = c(s_0,a_0) + \gamma \cdot c(s_1,a_1) + ...+ \gamma^{t-1} \cdot c(s_{t-1},a_{t-1}) \)\\
    v. Identify unsafe actions i.e. ${A}_\text{unsafe} = \{a_t \in {A}_\text{feasible} : C_{path} + \gamma^t \cdot \hat{Q}_{sc}^{*}(s_t,a_t) > \hat{c}\}$.\\
    vi. Expand tree for branches with safe actions, $a_t \in A \setminus A_\text{unsafe}$.\\
    
    $\hat{V}_R \gets \textsc{Rollout}(\mathcal{N}_{leaf})$ \tcp{SIMULATION (Get Monte Carlo reward estimate)}
    
    $\textsc{Backup}(\hat{V}_R,\mathcal{P})$ \tcp{BACKPROPAGATION (Update tree statistics)}
  }
\end{algorithm}

\textbf{Guided Bootstrapping of the Safety Critic Ensemble.} Training an ensemble for the safety critic is only half part of the story -- we still need informative data. A seamingly straight-forward approach solves the problem in Eq.~\ref{eq:constr_opt} in the offline phase, finds the optimal value $\lambda^*$ for the Lagrange multiplier, and utilizes the optimal, safe policy discovered to collect training data for the safety critic. In this case though, there is always the risk that the resulting safety critic (thus also the~\ac{MCTS} policy that utilizes it) does not generalize well far from the collected training data~\citep{ross2011reduction}.

Ideally, the training data covers the entire state-action space, but with a higher focus on states where selecting a specific action (over others) has a high effect on expected future performance~\citep{rexakis2012directed,kumar2022should} or cost violations/feasibility in our case. In other words: we must ensure that cost-critical states (i.e., states that -- in expectation under following the current policy $\pi$ -- have a high chance of violating the cost constraints later in the trajectories) are part of our training data. Our key idea is that we re-use the data that we collect during the optimizing of $\lambda$ during Lagrangian relaxation: We collect trajectories under different safety levels (i.e., different $\lambda$'s) which \emph{likely ensures} that we collect all cost-critical states around the constraint-switching hypersurface. See Appendix~\ref{app:guided-bootstrapping} for more details.

The iterative process of data gathering, followed by the training of a new version of ensemble safety critic is repeated until a safety critic leading to a \emph{feasible solution} is produced (as evaluated in the last phase shown in Fig.~\ref{fig:train_part}). See Appendix~\ref{app:considerations-reliability} for more details on the reliability of the safety critic.

\section{Evaluation}
\label{sec:evaluation}

We test our method by comparing its performance with the strongest -- to our knowledge -- baseline \ac{CC-MCP}~\citep{CC-MCP} on \emph{Rocksample} and \emph{Safe Gridworld} environments (see Sec.~\ref{supp:subsection:envs}). Considering that the scalability of MCTS has already been addressed in previous work (e.g,~\cite{silver2016mastering,schrittwieser2020mastering}), we have tried to define environments that are computationally manageable while still providing insights on properties and the quality of the final, feasible solution of our algorithm, w.r.t. the constraint formulation. See Appendix~\ref{supp:subsection:training_compute} for more details on the training setup and compute.

\begin{figure}[t!]
\begin{tikzpicture}
\begin{groupplot}[group style={group size=3 by 3, horizontal sep=1cm,vertical sep=1cm},
                   height=3.7cm, width=4.5cm,
                   xmin=128,xmax=8192,
                   ymin=0,
                   ymax=18,
                   grid=major,
                   ylabel style={align=center, text width=4cm},
                   minor y tick num=4,
                   minor x tick num=4,
                   tick label style={font=\footnotesize},
                   axis x line=bottom, axis y line=left, tick align=outside,
                   legend columns=-1,
                   legend style={/tikz/every even column/.append style={column sep=0.1cm},at={(0.5,1)},anchor=south,draw=none}, xmode = log, legend to name=legend,log basis x=2%
]
\nextgroupplot[title=\textit{Rocksample(5,7)},ymin=5,ymax=17.5,ylabel={Average Discounted Cumulative Reward}]
\addplot[draw=red, mark=o] table[draw=red, x={Number_of_Planning_Iterations},y={Average_Discounted_Cumulative_Reward}] {results/rocksample_5_7_ccmcp.dat};
\addplot[draw=blue, mark=+] table[x={Number_of_Planning_Iterations},y={Average_Discounted_Cumulative_Reward}] {results/rocksample_5_7_mctslambda.dat};
\addplot[draw=green,mark=triangle] table[x={Number_of_Planning_Iterations},y={Average_Discounted_Cumulative_Reward}] {results/rocksample_5_7_safemcts.dat};

\addplot [name path=upper1,draw=none] table[x=Number_of_Planning_Iterations,y expr=\thisrow{Average_Discounted_Cumulative_Reward}+\thisrow{Reward_Standard_Error}] {results/rocksample_5_7_ccmcp.dat};
\addplot [name path=lower1,draw=none] table[x=Number_of_Planning_Iterations,y expr=\thisrow{Average_Discounted_Cumulative_Reward}-\thisrow{Reward_Standard_Error}] {results/rocksample_5_7_ccmcp.dat};
\addplot [draw=red, fill=red!10] fill between[of=upper1 and lower1];

\addplot [name path=upper2,draw=none] table[x=Number_of_Planning_Iterations,y expr=\thisrow{Average_Discounted_Cumulative_Reward}+\thisrow{Reward_Standard_Error}] {results/rocksample_5_7_mctslambda.dat};
\addplot [name path=lower2,draw=none] table[x=Number_of_Planning_Iterations,y expr=\thisrow{Average_Discounted_Cumulative_Reward}-\thisrow{Reward_Standard_Error}] {results/rocksample_5_7_mctslambda.dat};
\addplot [draw=blue, fill=blue!10] fill between[of=upper2 and lower2];

\addplot [name path=upper4,draw=none] table[x=Number_of_Planning_Iterations,y expr=\thisrow{Average_Discounted_Cumulative_Reward}+\thisrow{Reward_Standard_Error}] {results/rocksample_5_7_safemcts.dat};
\addplot [name path=lower4,draw=none] table[x=Number_of_Planning_Iterations,y expr=\thisrow{Average_Discounted_Cumulative_Reward}-\thisrow{Reward_Standard_Error}] {results/rocksample_5_7_safemcts.dat};
\addplot [draw=green, fill=green!10] fill between[of=upper4 and lower4];

\nextgroupplot[title=\textit{Rocksample(7,8)},ymin=5,ymax=17.5]
\addplot[draw=red, mark=o] table[draw=red, x={Number_of_Planning_Iterations},y={Average_Discounted_Cumulative_Reward}] {results/rocksample_7_8_ccmcp.dat};
\addplot[draw=blue, mark=+] table[x={Number_of_Planning_Iterations},y={Average_Discounted_Cumulative_Reward}] {results/rocksample_7_8_mctslambda.dat};
\addplot[draw=green,mark=triangle] table[x={Number_of_Planning_Iterations},y={Average_Discounted_Cumulative_Reward}] {results/rocksample_7_8_safemcts.dat};

\addplot [name path=upper1,draw=none] table[x=Number_of_Planning_Iterations,y expr=\thisrow{Average_Discounted_Cumulative_Reward}+\thisrow{Reward_Standard_Error}] {results/rocksample_7_8_ccmcp.dat};
\addplot [name path=lower1,draw=none] table[x=Number_of_Planning_Iterations,y expr=\thisrow{Average_Discounted_Cumulative_Reward}-\thisrow{Reward_Standard_Error}] {results/rocksample_7_8_ccmcp.dat};
\addplot [draw=red, fill=red!10] fill between[of=upper1 and lower1];

\addplot [name path=upper2,draw=none] table[x=Number_of_Planning_Iterations,y expr=\thisrow{Average_Discounted_Cumulative_Reward}+\thisrow{Reward_Standard_Error}] {results/rocksample_7_8_mctslambda.dat};
\addplot [name path=lower2,draw=none] table[x=Number_of_Planning_Iterations,y expr=\thisrow{Average_Discounted_Cumulative_Reward}-\thisrow{Reward_Standard_Error}] {results/rocksample_7_8_mctslambda.dat};
\addplot [draw=blue, fill=blue!10] fill between[of=upper2 and lower2];

\addplot [name path=upper4,draw=none] table[x=Number_of_Planning_Iterations,y expr=\thisrow{Average_Discounted_Cumulative_Reward}+\thisrow{Reward_Standard_Error}] {results/rocksample_7_8_safemcts.dat};
\addplot [name path=lower4,draw=none] table[x=Number_of_Planning_Iterations,y expr=\thisrow{Average_Discounted_Cumulative_Reward}-\thisrow{Reward_Standard_Error}] {results/rocksample_7_8_safemcts.dat};
\addplot [draw=green, fill=green!10] fill between[of=upper4 and lower4];

\nextgroupplot[title=\textit{Rocksample(11,11)},ymin=0,ymax=10]
\addplot[draw=red, mark=o] table[draw=red, x={Number_of_Planning_Iterations},y={Average_Discounted_Cumulative_Reward}] {results/rocksample_11_11_ccmcp.dat};
\addplot[draw=blue, mark=+] table[x={Number_of_Planning_Iterations},y={Average_Discounted_Cumulative_Reward}] {results/rocksample_11_11_mctslambda.dat};
\addplot[draw=green,mark=triangle] table[x={Number_of_Planning_Iterations},y={Average_Discounted_Cumulative_Reward}] {results/rocksample_11_11_safemcts.dat};

\addplot [name path=upper1,draw=none] table[x=Number_of_Planning_Iterations,y expr=\thisrow{Average_Discounted_Cumulative_Reward}+\thisrow{Reward_Standard_Error}] {results/rocksample_11_11_ccmcp.dat};
\addplot [name path=lower1,draw=none] table[x=Number_of_Planning_Iterations,y expr=\thisrow{Average_Discounted_Cumulative_Reward}-\thisrow{Reward_Standard_Error}] {results/rocksample_11_11_ccmcp.dat};
\addplot [draw=red, fill=red!10] fill between[of=upper1 and lower1];

\addplot [name path=upper2,draw=none] table[x=Number_of_Planning_Iterations,y expr=\thisrow{Average_Discounted_Cumulative_Reward}+\thisrow{Reward_Standard_Error}] {results/rocksample_11_11_mctslambda.dat};
\addplot [name path=lower2,draw=none] table[x=Number_of_Planning_Iterations,y expr=\thisrow{Average_Discounted_Cumulative_Reward}-\thisrow{Reward_Standard_Error}] {results/rocksample_11_11_mctslambda.dat};
\addplot [draw=blue, fill=blue!10] fill between[of=upper2 and lower2];

\addplot [name path=upper4,draw=none] table[x=Number_of_Planning_Iterations,y expr=\thisrow{Average_Discounted_Cumulative_Reward}+\thisrow{Reward_Standard_Error}] {results/rocksample_11_11_safemcts.dat};
\addplot [name path=lower4,draw=none] table[x=Number_of_Planning_Iterations,y expr=\thisrow{Average_Discounted_Cumulative_Reward}-\thisrow{Reward_Standard_Error}] {results/rocksample_11_11_safemcts.dat};
\addplot [draw=green, fill=green!10] fill between[of=upper4 and lower4];

\nextgroupplot[ymax=1,ylabel={Average Discounted Cumulative Cost}]
\addplot[draw=red, mark=o] table[draw=red, x={Number_of_Planning_Iterations},y={Average_Discounted_Cumulative_Cost}] {results/rocksample_5_7_ccmcp.dat};
\addlegendentry{CC-MCP}
\addplot[draw=blue, mark=+] table[x={Number_of_Planning_Iterations},y={Average_Discounted_Cumulative_Cost}] {results/rocksample_5_7_mctslambda.dat};
\addlegendentry{MCTS $(\lambda=0.7)$}
\addplot[draw=green,mark=triangle] table[x={Number_of_Planning_Iterations},y={Average_Discounted_Cumulative_Cost}] {results/rocksample_5_7_safemcts.dat};
\addlegendentry{C-MCTS}

\addplot [name path=upper1,draw=none] table[x=Number_of_Planning_Iterations,y expr=\thisrow{Average_Discounted_Cumulative_Cost}+\thisrow{Cost_Standard_Error}] {results/rocksample_5_7_ccmcp.dat};
\addplot [name path=lower1,draw=none] table[x=Number_of_Planning_Iterations,y expr=\thisrow{Average_Discounted_Cumulative_Cost}-\thisrow{Cost_Standard_Error}] {results/rocksample_5_7_ccmcp.dat};
\addplot [draw=red, fill=red!10] fill between[of=upper1 and lower1];

\addplot [name path=upper2,draw=none] table[x=Number_of_Planning_Iterations,y expr=\thisrow{Average_Discounted_Cumulative_Cost}+\thisrow{Cost_Standard_Error}] {results/rocksample_5_7_mctslambda.dat};
\addplot [name path=lower2,draw=none] table[x=Number_of_Planning_Iterations,y expr=\thisrow{Average_Discounted_Cumulative_Cost}-\thisrow{Cost_Standard_Error}] {results/rocksample_5_7_mctslambda.dat};
\addplot [draw=blue, fill=blue!10] fill between[of=upper2 and lower2];

\addplot [name path=upper4,draw=none] table[x=Number_of_Planning_Iterations,y expr=\thisrow{Average_Discounted_Cumulative_Cost}+\thisrow{Cost_Standard_Error}] {results/rocksample_5_7_safemcts.dat};
\addplot [name path=lower4,draw=none] table[x=Number_of_Planning_Iterations,y expr=\thisrow{Average_Discounted_Cumulative_Cost}-\thisrow{Cost_Standard_Error}] {results/rocksample_5_7_safemcts.dat};
\addplot [draw=green, fill=green!10] fill between[of=upper4 and lower4];

\node at (axis cs:300,0.9) {$\hat{c}=1$};

\nextgroupplot[ymax=1]
\addplot[draw=red, mark=o] table[draw=red, x={Number_of_Planning_Iterations},y={Average_Discounted_Cumulative_Cost}] {results/rocksample_7_8_ccmcp.dat};
\addplot[draw=blue, mark=+] table[x={Number_of_Planning_Iterations},y={Average_Discounted_Cumulative_Cost}] {results/rocksample_7_8_mctslambda.dat};
\addplot[draw=green,mark=triangle] table[x={Number_of_Planning_Iterations},y={Average_Discounted_Cumulative_Cost}] {results/rocksample_7_8_safemcts.dat};

\addplot [name path=upper1,draw=none] table[x=Number_of_Planning_Iterations,y expr=\thisrow{Average_Discounted_Cumulative_Cost}+\thisrow{Cost_Standard_Error}] {results/rocksample_7_8_ccmcp.dat};
\addplot [name path=lower1,draw=none] table[x=Number_of_Planning_Iterations,y expr=\thisrow{Average_Discounted_Cumulative_Cost}-\thisrow{Cost_Standard_Error}] {results/rocksample_7_8_ccmcp.dat};
\addplot [draw=red, fill=red!10] fill between[of=upper1 and lower1];

\addplot [name path=upper2,draw=none] table[x=Number_of_Planning_Iterations,y expr=\thisrow{Average_Discounted_Cumulative_Cost}+\thisrow{Cost_Standard_Error}] {results/rocksample_7_8_mctslambda.dat};
\addplot [name path=lower2,draw=none] table[x=Number_of_Planning_Iterations,y expr=\thisrow{Average_Discounted_Cumulative_Cost}-\thisrow{Cost_Standard_Error}] {results/rocksample_7_8_mctslambda.dat};
\addplot [draw=blue, fill=blue!10] fill between[of=upper2 and lower2];

\addplot [name path=upper4,draw=none] table[x=Number_of_Planning_Iterations,y expr=\thisrow{Average_Discounted_Cumulative_Cost}+\thisrow{Cost_Standard_Error}] {results/rocksample_7_8_safemcts.dat};
\addplot [name path=lower4,draw=none] table[x=Number_of_Planning_Iterations,y expr=\thisrow{Average_Discounted_Cumulative_Cost}-\thisrow{Cost_Standard_Error}] {results/rocksample_7_8_safemcts.dat};
\addplot [draw=green, fill=green!10] fill between[of=upper4 and lower4];

\node at (axis cs:300,0.9) {$\hat{c}=1$};

\nextgroupplot[ymax=1.1]
\addplot[draw=red, mark=o] table[draw=red, x={Number_of_Planning_Iterations},y={Average_Discounted_Cumulative_Cost}] {results/rocksample_11_11_ccmcp.dat};
\addplot[draw=blue, mark=+] table[x={Number_of_Planning_Iterations},y={Average_Discounted_Cumulative_Cost}] {results/rocksample_11_11_mctslambda.dat};
\addplot[draw=green,mark=triangle] table[x={Number_of_Planning_Iterations},y={Average_Discounted_Cumulative_Cost}] {results/rocksample_11_11_safemcts.dat};

\addplot [name path=upper1,draw=none] table[x=Number_of_Planning_Iterations,y expr=\thisrow{Average_Discounted_Cumulative_Cost}+\thisrow{Cost_Standard_Error}] {results/rocksample_11_11_ccmcp.dat};
\addplot [name path=lower1,draw=none] table[x=Number_of_Planning_Iterations,y expr=\thisrow{Average_Discounted_Cumulative_Cost}-\thisrow{Cost_Standard_Error}] {results/rocksample_11_11_ccmcp.dat};
\addplot [draw=red, fill=red!10] fill between[of=upper1 and lower1];

\addplot [name path=upper2,draw=none] table[x=Number_of_Planning_Iterations,y expr=\thisrow{Average_Discounted_Cumulative_Cost}+\thisrow{Cost_Standard_Error}] {results/rocksample_11_11_mctslambda.dat};
\addplot [name path=lower2,draw=none] table[x=Number_of_Planning_Iterations,y expr=\thisrow{Average_Discounted_Cumulative_Cost}-\thisrow{Cost_Standard_Error}] {results/rocksample_11_11_mctslambda.dat};
\addplot [draw=blue, fill=blue!10] fill between[of=upper2 and lower2];

\addplot [name path=upper4,draw=none] table[x=Number_of_Planning_Iterations,y expr=\thisrow{Average_Discounted_Cumulative_Cost}+\thisrow{Cost_Standard_Error}] {results/rocksample_11_11_safemcts.dat};
\addplot [name path=lower4,draw=none] table[x=Number_of_Planning_Iterations,y expr=\thisrow{Average_Discounted_Cumulative_Cost}-\thisrow{Cost_Standard_Error}] {results/rocksample_11_11_safemcts.dat};
\addplot [draw=green, fill=green!10] fill between[of=upper4 and lower4];

\node at (axis cs:300,0.9) {$\hat{c}=1$};

\nextgroupplot[ymax=70,ylabel={Episodes With Cost Violations [\%]},xlabel={Planning Iterations}]
\addplot[draw=red, mark=o] table[draw=red, x={Number_of_Planning_Iterations},y={Number_of_Cost_Violations}] {results/rocksample_5_7_ccmcp.dat};
\addlegendentry{CC-MCP}
\addplot[draw=blue, mark=+] table[x={Number_of_Planning_Iterations},y={Number_of_Cost_Violations}] {results/rocksample_5_7_mctslambda.dat};
\addlegendentry{MCTS $(\lambda=0.7)$}
\addplot[draw=green,mark=triangle] table[x={Number_of_Planning_Iterations},y={Number_of_Cost_Violations}] {results/rocksample_5_7_safemcts.dat};
\addlegendentry{C-MCTS}

\nextgroupplot[ymax=70,xlabel={Planning Iterations}]
\addplot[draw=red, mark=o] table[draw=red, x={Number_of_Planning_Iterations},y={Number_of_Cost_Violations}] {results/rocksample_7_8_ccmcp.dat};
\addplot[draw=blue, mark=+] table[x={Number_of_Planning_Iterations},y={Number_of_Cost_Violations}] {results/rocksample_7_8_mctslambda.dat};
\addplot[draw=green,mark=triangle] table[x={Number_of_Planning_Iterations},y={Number_of_Cost_Violations}] {results/rocksample_7_8_safemcts.dat};

\nextgroupplot[xlabel={Planning Iterations},ymax=85]
\addplot[draw=red, mark=o] table[draw=red, x={Number_of_Planning_Iterations},y={Number_of_Cost_Violations}] {results/rocksample_11_11_ccmcp.dat};
\addlegendentry{CC-MCP}
\addplot[draw=blue, mark=+] table[x={Number_of_Planning_Iterations},y={Number_of_Cost_Violations}] {results/rocksample_11_11_mctslambda.dat};
\addlegendentry{MCTS}
\addplot[draw=green,mark=triangle] table[x={Number_of_Planning_Iterations},y={Number_of_Cost_Violations}] {results/rocksample_11_11_safemcts.dat};
\addlegendentry{C-MCTS}
\end{groupplot}
\node at (5.6,-7.8) {\pgfplotslegendfromname{legend}};
\end{tikzpicture}
\vspace{-4mm}
\caption{Performance of C-MCTS, MCTS, and CC-MCP on different \emph{Rocksample} configurations evaluated on 100 episodes. The shaded region represents the standard deviation over all episodes.}
\label{fig:rocksample_results}
\end{figure}

We evaluate the agent on different sizes and complexities of \emph{Rocksample} environments, with C-MCTS, \ac{CC-MCP}, and vanilla \ac{MCTS} (for penalized reward function with known $\lambda^*$). C-MCTS obtains higher rewards than \ac{CC-MCP} (see Fig.~\ref{fig:rocksample_results}, top row). The reward for C-MCTS increases with the number of planning iterations and the agent operates consistently below the cost-constraint (see Fig.~\ref{fig:rocksample_results}, middle row), close to the safety boundary. In constrast, \ac{CC-MCP} acts conservatively w.r.t costs and performs sub-optimally w.r.t. rewards as costs incurred in each episode vary greatly with different environment initializations. This is mitigated with C-MCTS since cost estimates with \ac{TD} learning have a lower variance than Monte Carlo cost estimates. Hence, the total number of cost violations is lower for C-MCTS compared to the other methods, in spite of operating closest to the safety constraint (see Fig.~\ref{fig:rocksample_results}, bottom row). Vanilla \ac{MCTS} obtains higher rewards than \ac{CC-MCP} as $\lambda^*$ is known, and unlike \ac{CC-MCP}, doesn't require tuning. \ac{MCTS} operates close to the cost-constraint but has a high number of cost violations. Compared to vanilla \ac{MCTS}, C-MCTS is safer, obtains equally high rewards, and in some cases even acts better (e.g., Rocksample($11,11$)). We refer the reader to Appendix~\ref{app:detailed_experiments} for a more detailed results on the planning efficiency of C-MCTS compared to CC-MCP (Appendix~\ref{app:planning_cost}).

\begin{figure}[t!]
    \centering
    \begin{tikzpicture}
    \begin{groupplot}[group style={group size=3 by 2, horizontal sep=1cm,vertical sep=1.5cm},
                   height=4cm, width=4.5cm,
                   xmin=128,xmax=8192,
                   ymin=0,
                   ymax=15,
                    ylabel style={align=center, text width=4cm},
                    title style={align=center, text width=4.43cm},
                   grid=major,
                   minor y tick num=4,
                   minor x tick num=4,
                   tick label style={font=\footnotesize},
                   axis x line=bottom, axis y line=left, tick align=outside,
                   legend columns=-1,
              legend style={/tikz/every even column/.append style={column sep=0.1cm},at={(0.5,1)},anchor=south,yshift=-5mm,draw=none}, xmode = log,log basis x=2%
    ]
    \nextgroupplot[ylabel=Average Discounted Cumulative Cost,xlabel=Planning Iterations,ymax=2,ymin=0.75,title=\textit{Planning Horizon}]
    \addplot[draw=orange, mark=o] table[draw=orange, x={Number_of_Planning_Iterations},y={Average_Discounted_Cumulative_Cost}] {results/ablation_study/planning_horizon_itr128_alpha4_eps0.1_sigmamax0.5.dat};

    \addplot[draw=cyan,mark=triangle] table[x={Number_of_Planning_Iterations},y={Average_Discounted_Cumulative_Cost}] {results/ablation_study/planning_horizon_itr1024_alpha4_eps0.1_sigmamax0.5.dat};
    \nextgroupplot[ymax=1.5,ymin=0.75,title=\textit{Ensemble Threshold},xlabel=Planning Iterations]
    \addplot[draw=gray, mark=pentagon] table[draw=orange, x={Number_of_Planning_Iterations},y={Average_Discounted_Cumulative_Cost}] {results/ablation_study/ensemble_threshold_itr512_alpha8_eps0.3_sigmamax0.1.dat};

    \addplot[draw=brown, mark=square] table[x={Number_of_Planning_Iterations},y={Average_Discounted_Cumulative_Cost}] {results/ablation_study/ensemble_threshold_itr512_alpha8_eps0.3_sigmamax0.5.dat};
         \nextgroupplot[ymax=2.6,ymin=0.75,title=\textit{Simulator Accuracy},xlabel=Planning Iterations]
    \addplot[draw=black, mark=otimes] table[draw=orange, x={Number_of_Planning_Iterations},y={Average_Discounted_Cumulative_Cost}] {results/ablation_study/imperfect_sim_alpha1_eps0.1_sigmamax0.5_d060.dat};

    \addplot[draw=purple, mark=diamond] table[x={Number_of_Planning_Iterations},y={Average_Discounted_Cumulative_Cost}] {results/ablation_study/imperfect_sim_alpha1_eps0.1_sigmamax0.5_d030.dat};
    
    
    \nextgroupplot[ylabel={Episodes With Cost Violations [\%]},ymax=18,legend to name=legend_training_horizon]
    \addplot[draw=orange, mark=o] table[draw=orange, x={Number_of_Planning_Iterations},y={Number_of_Cost_Violations}] {results/ablation_study/planning_horizon_itr128_alpha4_eps0.1_sigmamax0.5.dat};

    \addplot[draw=cyan,mark=triangle] table[x={Number_of_Planning_Iterations},y={Number_of_Cost_Violations}] {results/ablation_study/planning_horizon_itr1024_alpha4_eps0.1_sigmamax0.5.dat};
    \addlegendentry{128 planning iterations (training)}
    \addlegendentry{1024 planning iterations (training)}
    \nextgroupplot[ymax=20,legend to name=legend_ensemble]
    \addplot[draw=gray, mark=pentagon] table[draw=orange, x={Number_of_Planning_Iterations},y={Number_of_Cost_Violations}] {results/ablation_study/ensemble_threshold_itr512_alpha8_eps0.3_sigmamax0.1.dat};

    \addplot[draw=brown, mark=square] table[x={Number_of_Planning_Iterations},y={Number_of_Cost_Violations}] {results/ablation_study/ensemble_threshold_itr512_alpha8_eps0.3_sigmamax0.5.dat};

    \addlegendentry{$\sigma_{max}=0.1$}
    \addlegendentry{$\sigma_{max}=0.5$}

    \nextgroupplot[ymax=30,legend to name=legend_simulator]
     \addplot[draw=black, mark=otimes] table[draw=black, x={Number_of_Planning_Iterations},y={Number_of_Cost_Violations}] {results/ablation_study/imperfect_sim_alpha1_eps0.1_sigmamax0.5_d060.dat};
    \addplot[draw=purple, mark=diamond] table[x={Number_of_Planning_Iterations},y={Number_of_Cost_Violations}] {results/ablation_study/imperfect_sim_alpha1_eps0.1_sigmamax0.5_d030.dat};



    \addlegendentry{$\Delta d_0=40$}
    \addlegendentry{$\Delta d_0=10$}
    \end{groupplot}
        \node at (5.6,-5.1) {\pgfplotslegendfromname{legend_training_horizon}};
        \node at (1.6,-5.8) {\pgfplotslegendfromname{legend_ensemble}};
        \node at (7.6,-5.8) {\pgfplotslegendfromname{legend_simulator}};

    \end{tikzpicture}
    \vspace{-4mm}
    \caption{Comparing safety for different training/deployment strategies, i.e., using different planning horizons during training (left), deploying with different ensemble thresholds (middle), and collecting training samples from simulators of different accuracies (right).}
    \label{fig:planning_horizon_training}
\end{figure}

\textbf{Hyperparameters.} We optimized algorithmic parameters, i.e., $\alpha_0$ (initial step size to update $\lambda$) and $\epsilon$ (termination criterion for training loop) with a grid search (see values below), and ablated the remaining hyper-parameters hyperparameters, i.e., planning horizon in training, standard deviation threshold of the ensemble $\sigma$, and reliability of the simulator (controlled by $d_0$, see Sec.~\ref{supp:subsubsection:rocksample}). We conducted these experiments on Rocksample($7,8$) and averaged the results over 100 runs. 

\textit{Length of planning horizon during training (for $\alpha_0$=$4$ and $\epsilon$=$0.1$).} Regarding the effect of the planning horizon in the quality of the final solution, Fig.~\ref{fig:planning_horizon_training} (left column) indicates that the safety critic trained with a longer planning horizon operates closer to the safety boundary. This is because the safety critic predicts costs for a near-optimal policy and hence discerns the safety boundary more accurately. The safety critic trained with a smaller planning horizon estimates costs from a sub-optimal policy leading to cost violations during deployment. 

\textit{Ensemble threshold during deployment (for $\alpha_0$=$8$ and $\epsilon$=$0.3$).} We set different standard deviation thresholds ($\sigma_{max}=0.1$ and $\sigma_{max}=0.5$) in the neural network ensemble during deployment.
Fig.~\ref{fig:planning_horizon_training} (middle column) shows that the cost incurred exceeds the cost-constraint if $\sigma_{max}=0.1$, but the agent performs safely within the cost-constraint with a far lesser number of cost violations if $\sigma_{max}=0.5$. This is because we prune unsafe branches during planning only when the predictions between the individual members of the ensemble align with each other. Setting $\sigma_{max}=0.1$ is a tight bound resulting in most of the predictions of the safety critic being ignored. Using a higher threshold with $\sigma_{max}=0.5$ ensures that only large mismatches between the predictions of the individual members (corresponding to out-of-distribution inputs) are ignored, and the rest are used during planning. This results in the agent performing safely within the cost-constraint, but not too conservatively. 

\textit{Training on imperfect simulators (for $\alpha_0$=$1$ and $\epsilon$=$0.1$).} On the Rocksample environment, the sensor characteristics measuring the quality of the rock are defined by the constant $d_0$ (see Sec.~\ref{supp:subsubsection:rocksample}). We overestimate the sensor accuracy in our training simulator by choosing $d_0^{sim}$ with error $\Delta d_0$ and observe the safety of the agent in the real world when trained on simulators with different values of $\Delta d_0$. Fig.~\ref{fig:planning_horizon_training} (right column) shows the results. The values of $\Delta d_0$ set to 10 and 40 correspond to a maximum prediction error of $11.7\%$ and $32.5\%$, respectively. When $\Delta d_0=40$ the agent operates at a greater distance from the cost-constraint. The reason for cost violations is that the safety critic has been trained to place too much trust in the sensor measurements due to the simulation-to-reality gap. With a smaller gap ($\Delta d_0=10$) the agent performs safer. 


\section{Conclusion}
\label{sec:conclusion}

C-MCTS solves \acp{CMDP} by learning cost-estimates in a pre-training phase from simulated data and pruning unsafe branches of the search tree during deployment. Compared to previous work, C-MCTS does not need to tune a Lagrange multiplier online, which leads to better planning efficiency and higher rewards. In our experiments on Rocksample environments, C-MCTS achieved maximum rewards surpassing previous work for small, medium, and large-sized grids with increasing complexity, while maintaining safer performance. As cost is estimated from a lower variance \ac{TD} target the agent can operate close to the safety boundary with minimal constraint violations. C-MCTS is also suited for safety-critical applications that use approximate planning models for fast inference. Our Safe Gridworld results demonstrate that even with an approximate planning model, safety can be learned separately using a more realistic simulator, resulting in zero constraint violations and improved safety.

\textbf{Acknowledgments.} This work was supported by the Bavarian Ministry for Economic Affairs, Infrastructure, Transport and Technology through the Center for Analytics-Data-Applications (ADA-Center) within the framework of ``BAYERN DIGITAL II''.

\section*{Broader Impact}
\label{section:broader impact}

While C-MCTS mitigates the reliance on the planning model to meet cost constraints through pre-training in a high-fidelity simulator, there may still be sim-to-reality gaps when learning cost estimates. This introduces the possibility of encountering unforeseen consequences in real-world scenarios. In the context of using C-MCTS in a human-AI interaction task, if minority groups are not adequately represented in the training simulator, inaccurate cost estimates might lead to potential harm to humans. However, C-MCTS addresses these gaps more effectively than previous methods by leveraging a more relaxed computational budget during the training phase (fast inference is only required during deployment). This allows more accurate modeling of the real world to include rare edge scenarios. 

\bibliographystyle{plainnat}
\bibliography{references}

\clearpage
\appendix

\section*{Supplementary Material}

\section{Methodology}

\subsection{Guided Bootstrapping of the Safety Critic Ensemble}
\label{app:guided-bootstrapping}

Ideally, the training data covers the entire state-action space, but with a higher focus on states where selecting a specific action (over others) has a high effect on expected future performance~\citep{rexakis2012directed,kumar2022should} or cost violations/feasibility in our case.
\begin{definition}
A state $s$ is said to be cost-non-critical if
\begin{equation}
   \forall \mathit{a} \in A,\quad \operatorname*{min}_{\mathit{a}'} Q_c^{\pi} (s, \mathit{a}') \leq Q_c^{\pi} (s, \mathit{a}) \leq \hat{c} \quad \text{or}\quad
   \hat{c} \leq \operatorname*{min}_{\mathit{a}'} Q_c^{\pi} (s, \mathit{a}') \leq Q_c^{\pi} (s, \mathit{a})
\end{equation}
\label{def:critical_states}\vspace{-6mm}
\end{definition}
In other words, in cost-non-critical states, selecting any action under the applied policy $\pi$ does not lead (in expectation) to a change in the constraint/threshold violation (positive or negative).\footnote{Note that a similar discussion, under the concept of $\epsilon$-reducible datasets (or parts of datasets), also exists in safe/constrained offline reinforcement learning approaches~\citep{liu2023constrained}.} 
Even though having more training data from cost-critical states is desirable, these do not frequently occur in trajectories generated by any policy $\pi$ (see also the discussion in~\cite{kumar2022should}). 

The question now lies in how to sample data from cost-critical states. To achieve this, C-MCTS varies the value of the Lagrange parameter $\lambda$ in the offline phase to obtain different sets of trajectories with different safety levels (``data gathering'' in Fig.~\ref{fig:train_part}), and all this data is utilized for the training of the (ensemble) safety critic. 

The value of $\lambda$ is adapted following the standard training process in Lagrangian relaxation/augmentation settings~\citep{bertsekas2014constrained}. Here, training iterates between calculating a new value $\lambda_k$ in each $k$-th iteration of the data gathering loop and solving the $k$-th~\ac{MDP} (using~\ac{MCTS}) with the penalized reward function $r(s,a)-\lambda_k\, c(s,a)$. The latest (ensemble) safety critic is used to prune unsafe paths in \ac{MCTS}, as described in Algo.~\ref{alg:c_mcts}, thus pushing data collection to either safe or unexplored unsafe paths. This also implies that the action space of the $k$-th~\ac{MDP} will be different (more restricted) than the action space of the original~\ac{CMDP}, under the effect of the safety critic.

The new value for $\lambda_k$ in each iteration is $\lambda_k = \lambda_{k-1} + \frac{\alpha_0}{k} \left(V_{C}^{k, *}-\hat{c}\right)$, with $V_{C}^{k, *}$ being the optimal $V_C$ for the optimal policy (for the $k$-th~\ac{MDP}) with a fixed $\lambda_k$ at data gathering iteration $k$. The data gathering loop is terminated when $\hat{c} - \epsilon \leq V_{C}^{k, *} \leq \hat{c}$. Here, $\alpha_0$ and $\epsilon$ are tunable hyper-parameters. 

\begin{proposition}
    This iterative optimization process converges asymptotically to the optimal $\lambda^{*}$, in the $k$-th~\ac{MDP}.
\end{proposition}

\textit{Proof sketch.} Previous work~\citep{kocsis2006bandit,silver2016mastering} shows that the~\ac{MCTS} policy converges to the optimal policy as the number of simulations increases, meaning that in each iteration $k$ we are (asymptotically) guaranteed to find the optimal solution in the $k$-th~\ac{MDP}. Based on this, and on the fact that $\lambda$ is updated following the gradient direction of $V_{C}^{k, *}-\hat{c}$, convergence to the optimal $\lambda^{*}$ is achieved~\citep{CC-MCP,tessler2018reward,mankowitz2020robust}.

As ~\ac{MCTS} with upper confidence bounds converges asymptotically to the optimal policy~\citep{kocsis2006bandit}, usually a time- or computational budget-limit is used to terminate learning~\citep{silver2016mastering, alphazero}. As we are interested in a \emph{feasible} solution, we terminate the training process (search for $\lambda^*$ effectively) in the data gathering phase only when enough data has been gathered and the cost constraints are satisfied (see ``data gathering'' phase in Fig.~\ref{fig:train_part}).

Since the value of $\lambda_k$ is iteratively converging to $\lambda^*$ in each ``data gathering'' phase shown in Fig.~\ref{fig:train_part}, state-action pairs around the constraint-switching hypersurface are collected. The use of all available data (generated by different policies $\pi_k$ as a result of all values of $\lambda_k$) for the safety critic training (``model training'' phase in Fig.~\ref{fig:train_part}),  ensures that a large collection of state-action pairs from both critical and non-critical states is available.\footnote{With this data mixture we train the safety critic using $(s,a)$ samples that have different cost-targets (due to different $\lambda$'s), some of them over- or under-estimating the ``true'' cost. We could e.g. give higher weight to data from trajectories where the value of $\lambda$ was close to $\lambda^{*}$, but we observed that using an ensemble of safety critics (see Sec.~\ref{app:robustness}) combined with using the latest safety critic in each ``data gathering'' outer loop, leads to ``correct'' cost data being predominant and thus to a robust final safety critic, possibly at the cost of collecting more data.} 

\subsection{Considerations on the Reliability of the  Safety Critic}
\label{app:considerations-reliability}

The iterative process of data gathering, followed by the training of a new version of ensemble safety critic is repeated until a safety critic leading to a \emph{feasible solution} is produced (as evaluated in the last phase shown in Fig.~\ref{fig:train_part}).

\begin{proposition}
    Let $S_c \subseteq S$ be the set of cost-critical states (see Definition~\ref{def:critical_states}). Let $B = \{(s_c,a) | s_c \in S_c\ \text{and}\ a \in A\}$ be the set of all cost-critical-state and action pairs for a given~\ac{MDP}. Then, there exists $B_p \subseteq B$, a set of cost-critical-state and action pairs for which the trained safety critic would over-estimate the expected discounted cumulative cost, and $B_n \subseteq B$, a set of cost-critical-state and action pairs for which the trained safety critic would under-estimate it.
    Then, $B_p \cup B_n = B$ and $B_p \cap B_n = \varnothing$.
    \label{propose:planner_errors}
\end{proposition}

What Proposition~\ref{propose:planner_errors} indicates is that the trained safety critic will under-estimate or over-estimate the expected cost of~\emph{every} cost-critical-state and action pair defined in the underlying~\ac{MDP} of the~\emph{high-fidelity} simulator (except perhaps for trivial predictions, such as in close-to-terminal states of simple MDPs). This is both due to numerical precision issues, as well as due to the utilization of the low-fidelity simulator in the~\ac{MCTS} planner, which potentially predicts sequences of safe or unsafe next states that are different compared to the actual ones, especially for cost-critical-state and action pairs that are far from the terminal states.

\begin{corollary}
    The overall training process of the safety critic, illustrated in Fig.~\ref{fig:train_part}, converges to a feasible solution of the constrained optimization problem defined in (\ref{eq:constr_opt}).
\end{corollary}

\textit{Proof sketch.} As discussed before, the inner training loop will asymptotically converge to the optimal solution in the $k-$th~\ac{MDP} (``data gathering'' phase in Fig.~\ref{fig:train_part}). In case the safety critic over-estimates the expected cost ($(s_c, a) \in B_p$), it will lead to pruning the corresponding branch in the~\ac{MCTS} tree. This leads to a \emph{safe}, but potentially \emph{conservative} (i.e., non-optimal) behavior. In case of under-estimation ($(s_c, a) \in B_n$), the respective branch can be traversed and a non-safe trajectory is performed at the high-fidelity simulator. Since data collected from the unsafe trajectories are used in subsequent safety critic training iterations, the new versions of the safety critic will no longer under-estimate the cost. This means, that progressively all the $(s_c, a) \in B_n$ pairs (as defined in Proposition~\ref{propose:planner_errors}) that are visited in the high-fidelity simulator will belong to the $B_p$ set in subsequent iterations and there will be no constraint violations eventually, i.e., we will have a \emph{feasible solution}.

\section{Details on the Experimental Setup}

\subsection{Environments}
\label{supp:subsection:envs}

\subsubsection{Rocksample}
\label{supp:subsubsection:rocksample}

\begin{figure}[t!]
    \centering
    \includegraphics[width=0.3\textwidth]{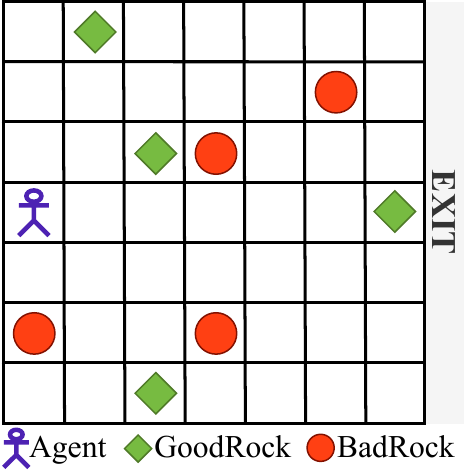}\hspace{2cm}
    \includegraphics[width=0.3\textwidth]{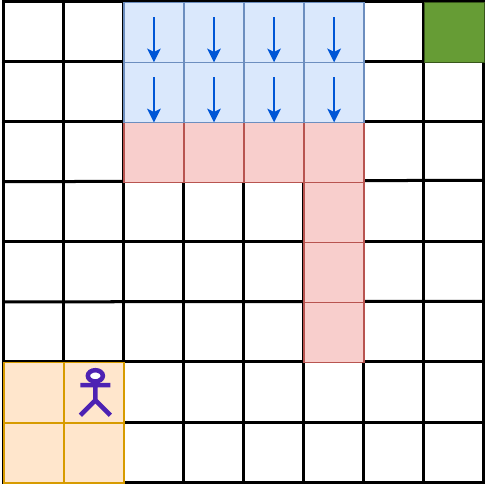}
    \caption{Environments: (left) exemplary Rocksample$(7,8)$ environment, i.e., a $7 \times 7$ rocksample environment with $8$ rocks randomly placed; (right) exemplary Safe Gridworld environment, where the colors denote start cells (yellow), the goal cell (green), unsafe cells (pink), and windy cells (blue).}
    \label{fig:envs}
\end{figure}

The environment is defined as a grid with $n \times n$ squares with $m$ rocks randomly placed, some being good and others bad (see Fig.~\ref{fig:envs}, left). A specific Rocksample setup is defined by the nomenclature Rocksample($n,m$). A rover (agent) starting from the left is tasked to collect as many good rocks as possible and exit the grid to the right. The positions of the rocks are known in advance, but the quality of the rocks is unknown.  The agent can move up, down, right, and left, sample a rock, or make measurements to sense the quality of a rock. The total number of possible actions is hence $5+m$. The agent is equipped with a noisy sensor to measure the quality of a rock with a probability of accuracy $(2^{-d/d_0} + 1)/2$, where $d$ is the Euclidean distance of the agent from the corresponding rock and $d_0$ is a constant. The number of measurements that the agent can perform is constrained. Trying to maximize rewards (collecting good rocks) with constraints (number of sensor measurements) encourages the agent to use a limited number of measurements at a reasonable proximity to the rocks, wherein the sensor readings can be trusted. At each time step the agent observes its own position and the positions of the rocks with the updated probabilities.

We formulate the task within the \ac{CMDP} framework by additionally defining a reward structure, cost function, and a cost-constraint. The agent is rewarded a +10 reward for exiting the grid from the right or for collecting a good rock. A -10 penalty is received for each bad rock collected, and a -100 penalty is given when the agent exits the grid to the other sides or if the agent tries to sample a rock from an empty grid location. The agent incurs a +1 cost when measuring the quality of a single rock. The discounted cost over an episode cannot exceed 1, and this is the cost-constraint. The discount factor $\gamma$ is set to 0.95.

\subsubsection{Safe Gridworld}
\label{supp:subsubsection:safegridworld}

We additionally propose a new problem: \emph{Safe Gridworld}. The environment is defined as $8 \times 8$ grid where an agent from the bottom left region is tasked to find the shortest path to reach the top right square avoiding unsafe squares on the way (see Fig.~\ref{fig:envs}, right). The agent can move to the neighboring squares and has a total of $9$ action choices. The transition dynamics in all squares are deterministic except the $8$ squares at the top which are stochastic. These squares have winds blowing from the top to the bottom forcefully pushing the agent down by one square with a probability of $0.3$, independent of the action chosen by the agent (we vary this probability to account for simulator mismatch in the experiment in Sec.~\ref{subsection:eval-sim2real}). Otherwise, the transition is guided by the agent's action.

The agent receives a reward of +100 on reaching the goal state, a -1000 penalty for exiting the grid, and a -1 penalty otherwise until the terminal state is reached. Entering an unsafe square incurs a cost of +1. The agent should only traverse safe squares, and the discounted cost over an episode is 0. The cost-constraint imposes this as a constraint and is set to 0. The discount factor $\gamma$ is set to 0.95.

\subsection{Training Details \& Compute}
\label{supp:subsection:training_compute}

The training and evaluation were conducted on a single Intel Xeon E3-1240 v6 CPU. The CPU specifications are listed below. 
\begin{table}[h]
\centering
\caption{Specifications of Intel Xeon E3-1240 v6}
\begin{tabular}{|c|c|}
\hline
\textbf{Component} & \textbf{Specification} \\
\hline
Generation & Kaby Lake \\
\hline
Number of Cores & 4 \\
\hline
Hyper-Threading (HT) & Disabled \\
\hline
Base Frequency & 3.70 GHz \\
\hline
RAM & 32 GB \\
\hline
SSD & 960 GB \\
\hline
\end{tabular}
\label{tab:cpu_specs}
\end{table}

No GPU accelerators were used as the C-MCTS implementation was not optimized for efficient GPU resource utilization. The hyperparameters chosen for training the safety critic in the primary results (Fig.~\ref{fig:rocksample_results}) are summarized in Table \ref{tab:hyperparamaters_training}.

\begin{table}[h]
\centering
\caption{Key hyperparameters to train the safety critic. }
\begin{tabular}{|l|c|c|c|c|}
\hline
\textbf{Environment} & $\boldsymbol{\alpha_0}$ & $\boldsymbol{\epsilon}$ & $\boldsymbol{\sigma_{max}}$ & \textbf{Planning Iterations} \\
\hline
Rocksample$(5,7)$ & 8  & 0.1 & 0.5 & 1024 \\
\hline
Rocksample$(7,8)$ & 4 & 0.1 & 0.5 & 1024 \\
\hline
Rocksample$(11,11)$ & 12 & 0.1 & 0.5 & 512 \\
\hline
Safe Gridworld  & 10 & 0.1 & 0.2 & 512 \\
\hline
\end{tabular}
\label{tab:hyperparamaters_training}
\end{table}

\clearpage

\section{Additional Experimental Results}
\label{app:detailed_experiments}

\subsection{Planning cost to achieve high rewards: C-MCTS vs \ac{CC-MCP}}
\label{app:planning_cost}
\begin{table}[ht]
\centering
\small 
$N/A$: Performance not achieved $\qquad \qquad$(*): Additional evaluation environment
\begin{tabular}{|c|c|>{\centering\arraybackslash}p{5cm}|>{\centering\arraybackslash}p{3cm}|}
\hline
Environment & Method & \multicolumn{2}{c|}{Performance} \\
\cline{3-4}
 & & Number of planning iterations & Discounted Reward \\
\hline
\multirow{2}{*}{Rocksample(5,7)} & CC-MCP & $2^{20}$ & 13.72 \\
 & C-MCTS & $2^{10}$ & 13.93 \\
\hline
\multirow{2}{*}{Rocksample(7,8)} & CC-MCP & $2^{20}$ & 9.83 \\
 & C-MCTS & $2^{10}$ & 11.0 \\
\hline
\multirow{2}{*}{Rocksample(11,11)} & CC-MCP & $2^{20}$ & 5.26 \\
 & C-MCTS & $2^{10}$ & 7.14 \\
 \hline
\multirow{2}{*}{Rocksample(15,15)$^*$} & CC-MCP & $N/A$ & $N/A$ \\
 & C-MCTS & $2^{8}$ & 14.29 \\
\hline
\end{tabular}
\caption{Comparing planning iterations of C-MCTS and CC-MCP at equivalent reward levels.}
\end{table}

\subsection{Computational cost comparison}
\label{app:comp_cost}

In terms of computational cost per simulation across the different algorithmic phases:
\begin{itemize}
    \item Selection: CC-MCP is the most computationally expensive, requiring more operations to select the best child node. MCTS and C-MCTS have identical operation counts.
    \item Expansion: C-MCTS incurs additional costs due to the safety critic's prediction. MCTS and CC-MCP require no additional computation during expansion.
    \item Backpropagation: CC-MCP backs up Q-values for both reward and cost, while MCTS and C-MCTS back up only Q-values for reward.
    \item Rollout: Computational cost is identical for C-MCTS, MCTS, and CC-MCP.
\end{itemize}

C-MCTS and MCTS algorithms were implemented in Python, while the benchmark CC-MCP uses a C++ implementation. Comparing actual execution times was unfair since our implementation was not optimized for hardware efficiency. Also, such an optimization would highly depend on the hardware platform (e.g. CPUs vs GPUs). For instance, added cost in C-MCTS's expansion phase is highly parallelizable, with good potential for effective GPU utilization. So, for our analysis we instead compare the number of simulations (planning iterations) required by each algorithm as a performance metric.

\subsection{Planning efficiency}
\label{app:planning_efficiency}

We compare the planning efficiency of all methods on the same set of experiments. The comparison is done based on the depth of the search tree, given a specific computational budget, i.e., a fixed number of planning iterations. This comparison is qualitative and is used to evaluate the effectiveness of different planning algorithms. Fig.~\ref{fig:peak_tree_depth} shows that C-MCTS performs a more narrow search for the same number of planning iterations. The peak tree depth (averaged over 100 episodes) is the highest for C-MCTS. In C-MCTS the exploration space is restricted by the safety critic, and this helps in efficient planning. In Rocksample($11,11$) the peak tree depth of \ac{CC-MCP} is high in spite of having a sub-optimal performance. This is probably because the Lagrange multiplier in \ac{CC-MCP} gets stuck in a local maximum and is unable to find the optimum.

\begin{figure}[t!]
    \centering
    \begin{tikzpicture}
    \begin{groupplot}[group style={group size=3 by 1, horizontal sep=1cm,vertical sep=1.5cm},
                   height=4cm, width=4.5cm,
                   xmin=128,xmax=8192,
                   ymin=0,
                   ymax=15,
                   xlabel=Planning Iterations,
                   grid=major,
                   minor y tick num=4,
                   minor x tick num=4,
                   ylabel style={align=center, text width=3.5cm},
                   tick label style={font=\footnotesize},
                   axis x line=bottom, axis y line=left, tick align=outside,
                   legend columns=-1,
                   legend style={/tikz/every even column/.append style={column sep=0.1cm},at={(0.5,1)},anchor=south,yshift=-5mm,draw=none}, xmode = log, legend to name=legend_tree,log basis x=2%
    ]
    \nextgroupplot[title=\textit{Rocksample(5,7)},ymax=8,ylabel= Average Peak Tree Depth]
    \addplot[draw=red, mark=o] table[draw=red, x={Number_of_Planning_Iterations},y={Tree_Depth}] {results/rocksample_5_7_ccmcp.dat};
    \addplot[draw=blue, mark=+] table[x={Number_of_Planning_Iterations},y={Tree_Depth}] {results/rocksample_5_7_mctslambda.dat};
    \addplot[draw=green,mark=triangle] table[x={Number_of_Planning_Iterations},y={Tree_Depth}] {results/rocksample_5_7_safemcts.dat};
    \addplot [name path=upper1,draw=none] table[x=Number_of_Planning_Iterations,y expr=\thisrow{Tree_Depth}+\thisrow{Tree_Depth_Error}] {results/rocksample_5_7_ccmcp.dat};
    \addplot [name path=lower1,draw=none] table[x=Number_of_Planning_Iterations,y expr=\thisrow{Tree_Depth}-\thisrow{Tree_Depth_Error}] {results/rocksample_5_7_ccmcp.dat};
    \addplot [draw=red, fill=red!10] fill between[of=upper1 and lower1];
    
    \addplot [name path=upper2,draw=none] table[x=Number_of_Planning_Iterations,y expr=\thisrow{Tree_Depth}+\thisrow{Tree_Depth_Error}] {results/rocksample_5_7_mctslambda.dat};
    \addplot [name path=lower2,draw=none] table[x=Number_of_Planning_Iterations,y expr=\thisrow{Tree_Depth}-\thisrow{Tree_Depth_Error}] {results/rocksample_5_7_mctslambda.dat};
    \addplot [draw=blue, fill=blue!10] fill between[of=upper2 and lower2];
    
    \addplot [name path=upper4,draw=none] table[x=Number_of_Planning_Iterations,y expr=\thisrow{Tree_Depth}+\thisrow{Tree_Depth_Error}] {results/rocksample_5_7_safemcts.dat};
    \addplot [name path=lower4,draw=none] table[x=Number_of_Planning_Iterations,y expr=\thisrow{Tree_Depth}-\thisrow{Tree_Depth_Error}] {results/rocksample_5_7_safemcts.dat};
    \addplot [draw=green, fill=green!10] fill between[of=upper4 and lower4];

    \nextgroupplot[title=\textit{Rocksample(7,8)},ymax=12]
    \addplot[draw=red, mark=o] table[draw=red, x={Number_of_Planning_Iterations},y={Tree_Depth}] {results/rocksample_7_8_ccmcp.dat};
    \addplot[draw=blue, mark=+] table[x={Number_of_Planning_Iterations},y={Tree_Depth}] {results/rocksample_7_8_mctslambda.dat};
    \addplot[draw=green,mark=triangle] table[x={Number_of_Planning_Iterations},y={Tree_Depth}] {results/rocksample_7_8_safemcts.dat};
    \addplot [name path=upper1,draw=none] table[x=Number_of_Planning_Iterations,y expr=\thisrow{Tree_Depth}+\thisrow{Tree_Depth_Error}] {results/rocksample_7_8_ccmcp.dat};
    \addplot [name path=lower1,draw=none] table[x=Number_of_Planning_Iterations,y expr=\thisrow{Tree_Depth}-\thisrow{Tree_Depth_Error}] {results/rocksample_7_8_ccmcp.dat};
    \addplot [draw=red, fill=red!10] fill between[of=upper1 and lower1];
    
    \addplot [name path=upper2,draw=none] table[x=Number_of_Planning_Iterations,y expr=\thisrow{Tree_Depth}+\thisrow{Tree_Depth_Error}] {results/rocksample_7_8_mctslambda.dat};
    \addplot [name path=lower2,draw=none] table[x=Number_of_Planning_Iterations,y expr=\thisrow{Tree_Depth}-\thisrow{Tree_Depth_Error}] {results/rocksample_7_8_mctslambda.dat};
    \addplot [draw=blue, fill=blue!10] fill between[of=upper2 and lower2];
    
    \addplot [name path=upper4,draw=none] table[x=Number_of_Planning_Iterations,y expr=\thisrow{Tree_Depth}+\thisrow{Tree_Depth_Error}] {results/rocksample_7_8_safemcts.dat};
    \addplot [name path=lower4,draw=none] table[x=Number_of_Planning_Iterations,y expr=\thisrow{Tree_Depth}-\thisrow{Tree_Depth_Error}] {results/rocksample_7_8_safemcts.dat};
    \addplot [draw=green, fill=green!10] fill between[of=upper4 and lower4];

     \nextgroupplot[title=\textit{Rocksample(11,11)},ymax=20]
    \addplot[draw=red, mark=o] table[draw=red, x={Number_of_Planning_Iterations},y={Tree_Depth}] {results/rocksample_11_11_ccmcp.dat};
    \addlegendentry{CC-MCP}
    \addplot[draw=blue, mark=+] table[x={Number_of_Planning_Iterations},y={Tree_Depth}] {results/rocksample_11_11_mctslambda.dat};
    \addlegendentry{MCTS}
    \addplot[draw=green,mark=triangle] table[x={Number_of_Planning_Iterations},y={Tree_Depth}] {results/rocksample_11_11_safemcts.dat};
    \addlegendentry{C-MCTS}
    \addplot [name path=upper1,draw=none] table[x=Number_of_Planning_Iterations,y expr=\thisrow{Tree_Depth}+\thisrow{Tree_Depth_Error}] {results/rocksample_11_11_ccmcp.dat};
    \addplot [name path=lower1,draw=none] table[x=Number_of_Planning_Iterations,y expr=\thisrow{Tree_Depth}-\thisrow{Tree_Depth_Error}] {results/rocksample_11_11_ccmcp.dat};
    \addplot [draw=red, fill=red!10] fill between[of=upper1 and lower1];
    
    \addplot [name path=upper2,draw=none] table[x=Number_of_Planning_Iterations,y expr=\thisrow{Tree_Depth}+\thisrow{Tree_Depth_Error}] {results/rocksample_11_11_mctslambda.dat};
    \addplot [name path=lower2,draw=none] table[x=Number_of_Planning_Iterations,y expr=\thisrow{Tree_Depth}-\thisrow{Tree_Depth_Error}] {results/rocksample_11_11_mctslambda.dat};
    \addplot [draw=blue, fill=blue!10] fill between[of=upper2 and lower2];
    
    \addplot [name path=upper4,draw=none] table[x=Number_of_Planning_Iterations,y expr=\thisrow{Tree_Depth}+\thisrow{Tree_Depth_Error}] {results/rocksample_11_11_safemcts.dat};
    \addplot [name path=lower4,draw=none] table[x=Number_of_Planning_Iterations,y expr=\thisrow{Tree_Depth}-\thisrow{Tree_Depth_Error}] {results/rocksample_11_11_safemcts.dat};
    \addplot [draw=green, fill=green!10] fill between[of=upper4 and lower4];
    
    \end{groupplot}
    \node at (5.6,-1.5) {\pgfplotslegendfromname{legend_tree}};
    \end{tikzpicture}
   
    \caption{Maximum depth of the search tree for C-MCTS, \ac{MCTS} and \ac{CC-MCP} on different rocksample configurations averaged over 100 episodes.}
    \label{fig:peak_tree_depth}
\end{figure}

\subsection{Robustness to source/target environment mismatch}
\label{app:robustness}

\textbf{MCTS planner model.}\label{subsection:eval-sim2real}  The benefit of learning safety constraints before deployment from a simulator that has a higher fidelity compared to the planning (low-fidelity) simulator is evident when examining the synthetically constructed \emph{Safe Gridworld} scenario (see Sec.~\ref{supp:subsubsection:safegridworld}). In this setup, we use a planning simulator that models the dynamics approximately, and a training simulator (for the safety critic) that captures the dynamics more accurately. In the planning simulator, all transition dynamics are accurately modeled, except the blue squares with winds (Fig.~\ref{fig:envs} right). The transitions here are determined by the action selection (stochasticity due to wind is not considered). The training simulator models the transitions in these regions more accurately, but with some errors. The agent in the blue squares moves down with a probability of $0.25$, as compared to the real-world configuration where the probability is $0.3$. 
We trained and evaluated C-MCTS for $2^9$ and \ac{CC-MCP} for $2^{20}$ planning iterations. The latter was set to a higher planning budget to allow the baseline algorithm to converge to its final solution.

\begin{figure}[t!]
    \centering
    \subfigure[C-MCTS with $0\%$ cost violations.]{
        \hspace*{1cm}
        \includegraphics[width=0.3\linewidth]{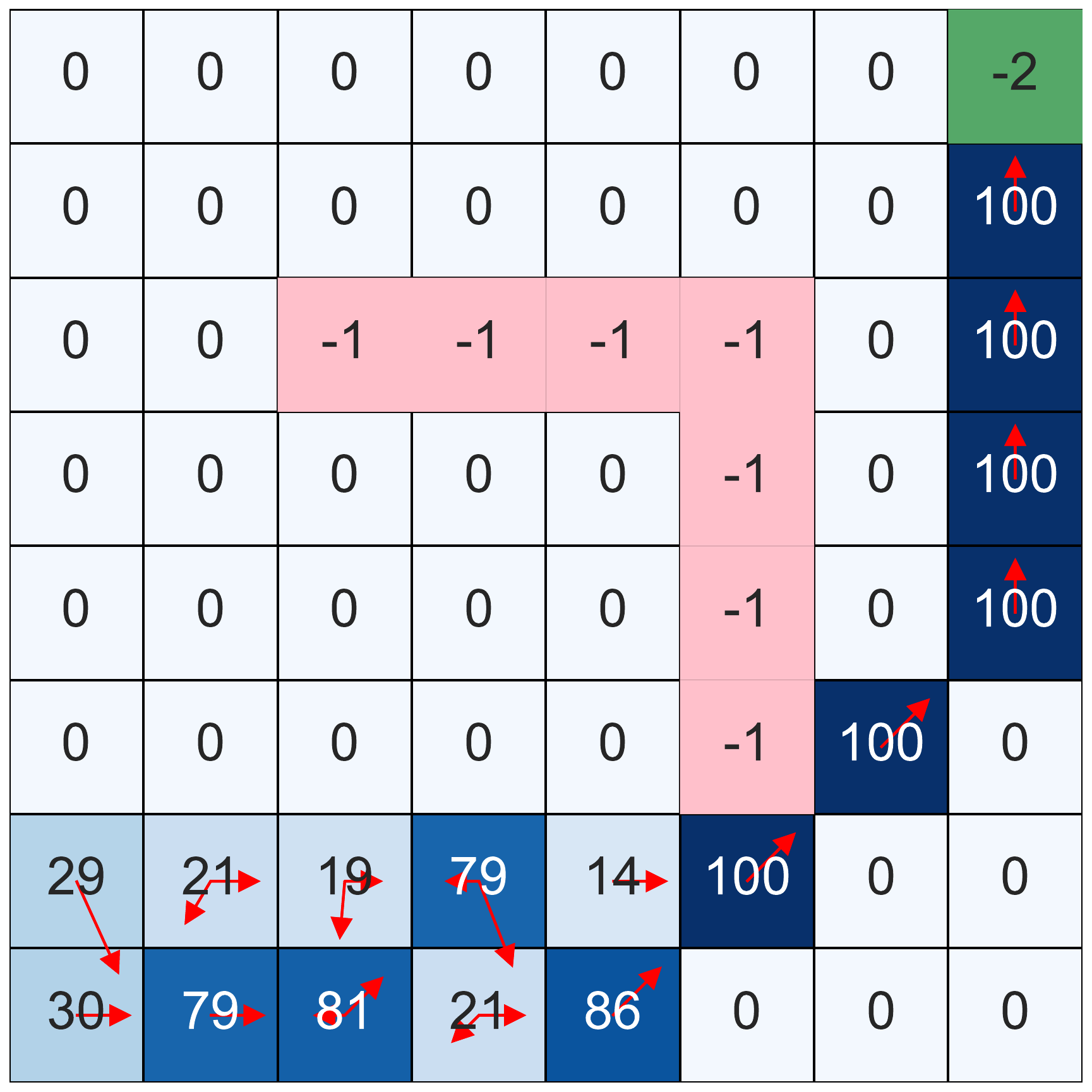}
        \hspace*{1cm}
    }
    \subfigure[\ac{CC-MCP} with $11\%$ cost violations.]{
        \hspace*{1cm}
        \includegraphics[width=0.3\linewidth]{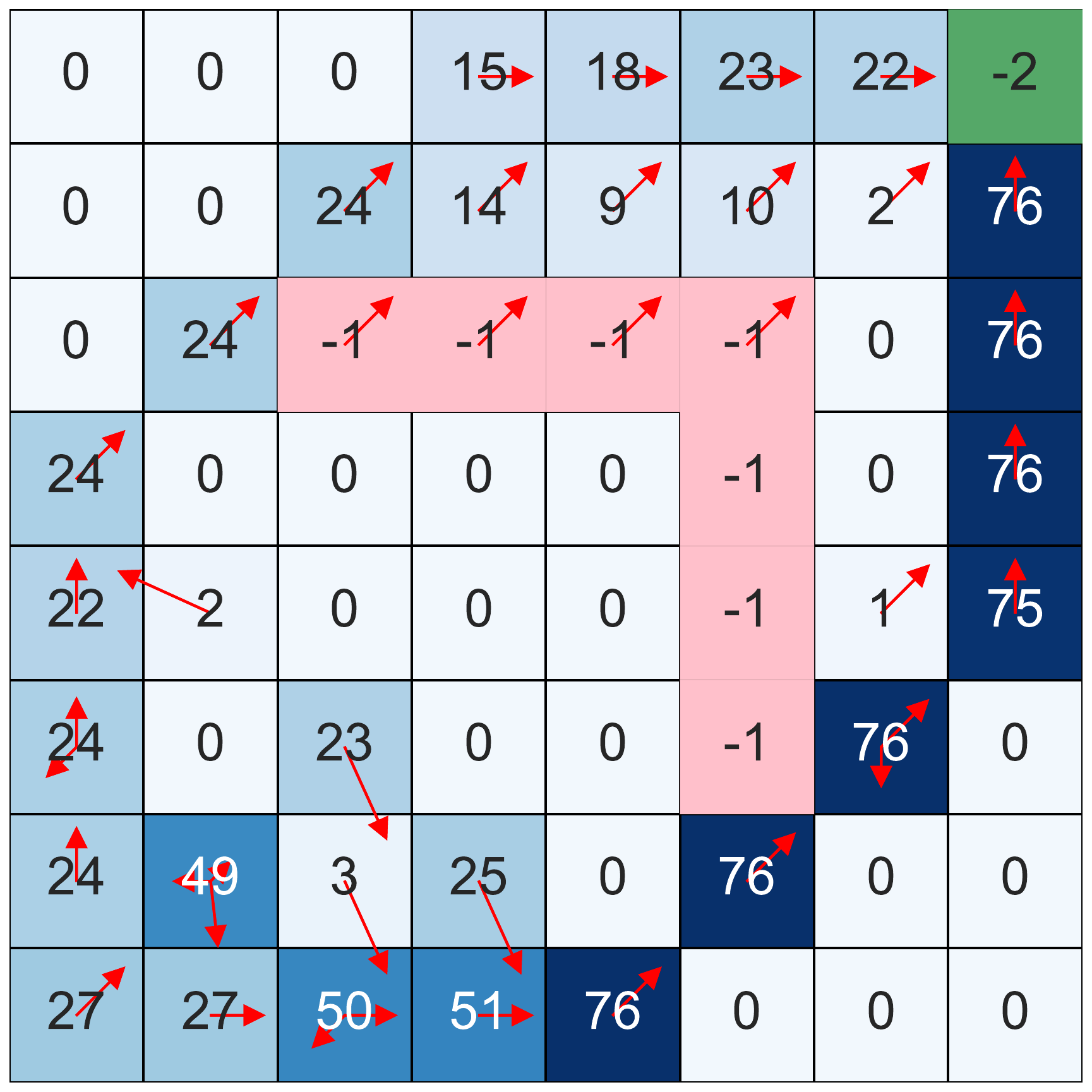}
        \hspace*{1cm}
    }
    \caption{State visitations aggregated over 100 episodes. The length of the arrows is proportional to the number of action selections. Values of -1 and -2 denote unsafe cells and the goal cell, respectively.}
    \label{fig:safegrid_results}
\end{figure}

Fig.~\ref{fig:safegrid_results} shows the number of state visitations of C-MCTS (left) and CC-MCP (right). The \ac{CC-MCP} agent takes both of the possible paths (going to the top and to the right), avoiding the unsafe region (in pink) to reach the goal state, which is optimal in the absence of the windy squares, but here it leads to cost violations due to inaccurate cost estimates.\footnote{Of course also the variance could play a minor role but we designed the setup to focus on the dynamics mismatch between the planner and the actual environment, which is much more prevalent here.} C-MCTS on the other hand only traverses through the two right-most columns to avoid the unsafe region, as the safety critic being trained using the high-fidelity simulator identifies the path from the top as unsafe, which leads to zero cost violations.

\end{document}